\documentclass[10pt,twocolumn,letterpaper]{article}

\usepackage{wacv}
\usepackage{times}
\usepackage{epsfig}
\usepackage{graphicx}
\usepackage{amsmath}
\usepackage{amssymb}

\usepackage[noend]{algpseudocode}
\usepackage{algorithm, algorithmicx}
\usepackage{subfigure, multirow}
\usepackage{algpseudocode}
\usepackage{booktabs}
\usepackage{array}
\usepackage{cite}
\usepackage{sidecap}
\usepackage[skip=3pt]{caption}
\usepackage[pagebackref=true,breaklinks=true,letterpaper=true,colorlinks,bookmarks=false]{hyperref}

\def\wacvPaperID{****} 

\wacvfinalcopy 

\ifwacvfinal
\def\assignedStartPage{1} 
\fi

\ifwacvfinal
\usepackage[breaklinks=true,bookmarks=false]{hyperref}
\else
\usepackage[pagebackref=true,breaklinks=true,colorlinks,bookmarks=false]{hyperref}
\fi

\ifwacvfinal
\else
\fi

\begin{document}

\title{Continual Representation Learning for Biometric Identification}

\author{Bo Zhao$^{*,1,2}$, Shixiang Tang$^{*,2,3}$, Dapeng Chen$^{2}$, Hakan Bilen$^1$, Rui Zhao$^2$\\
$^1$The University of Edinburgh, $^2$SenseTime Group Limited, $^3$The University of Sydney\\
$^*$Equal Contribution\\
{\tt\small \{bo.zhao, hbilen\}@ed.ac.uk, \{tangshixiang, chendapeng, zhaorui\}@sensetime.com}
}

\maketitle

\begin{abstract}
With the explosion of digital data in recent years, continuously learning new tasks from a stream of data  without forgetting previously acquired knowledge has become increasingly important. In this paper, we propose a new continual learning (CL) setting, namely ``continual representation learning'', which focuses on learning better representation in a continuous way. We also provide two large-scale multi-step benchmarks for biometric identification, where the visual appearance of different classes are highly relevant. In contrast to requiring the model to recognize more learned classes, we aim to learn feature representation that can be better generalized to not only previously unseen images but also unseen classes/identities. For the new setting, we propose a novel approach that performs the knowledge distillation over a large number of identities by applying the neighbourhood selection and consistency relaxation strategies to improve scalability and flexibility of the continual learning model. We demonstrate that existing CL methods can improve the representation in the new setting, and our method achieves better results than the competitors.
\end{abstract}

\section{Introduction}
Biometric identification \cite{jain2000biometric,sanchez2000biometric}, including face recognition \cite{liu2015deep,wen2016discriminative,deng2019arcface} and person re-identification (re-id) \cite{li2014deepreid,xu2018attention,wei2018person}, has achieved significant progress in the recent years due to the advances in modern learnable representations \cite{wen2016discriminative,deng2019arcface,li2014deepreid,schroff2015facenet,cheng2016person, hermans2017defense,varior2016gated,chen2017beyond} and emerging large datasets \cite{LFWTech,klare2015pushing,zheng2015scalable,guo2016ms,kemelmacher2016megaface,zheng2017unlabeled,wei2018person}. In particular, deep neural networks (DNN) \cite{simonyan2014very,ren2015faster,szegedy2015going,he2016deep} are shown to learn features that encode complex and mosaic biometrics traits and achieve better feature generalization ability, when trained on large-scale datasets. However, the paradigm of training DNNs offline becomes impractical and inefficient with the increase in stream data such as surveillance videos and online images/texts.
For example, the intelligent security system \cite{wang2013intelligent, zheng2016person} in a city or an airport captures millions of new images every day.  In this scenario, training a model with all the images in one step can never be realized.  To continuously improve our model with limited computational and storage resource, we expect the model to be trained online only with the newly obtained data.

Motivated by this, we propose a new but realistic setting named ``continual representation learning'' (CRL) for this real-world biometric identification problem. The new setting aims to learn from continuous stream data, meanwhile continuously improving model's generalization ability on unseen classes/identities.




\begin{figure}[t]
\centering
\includegraphics[width=1\linewidth]{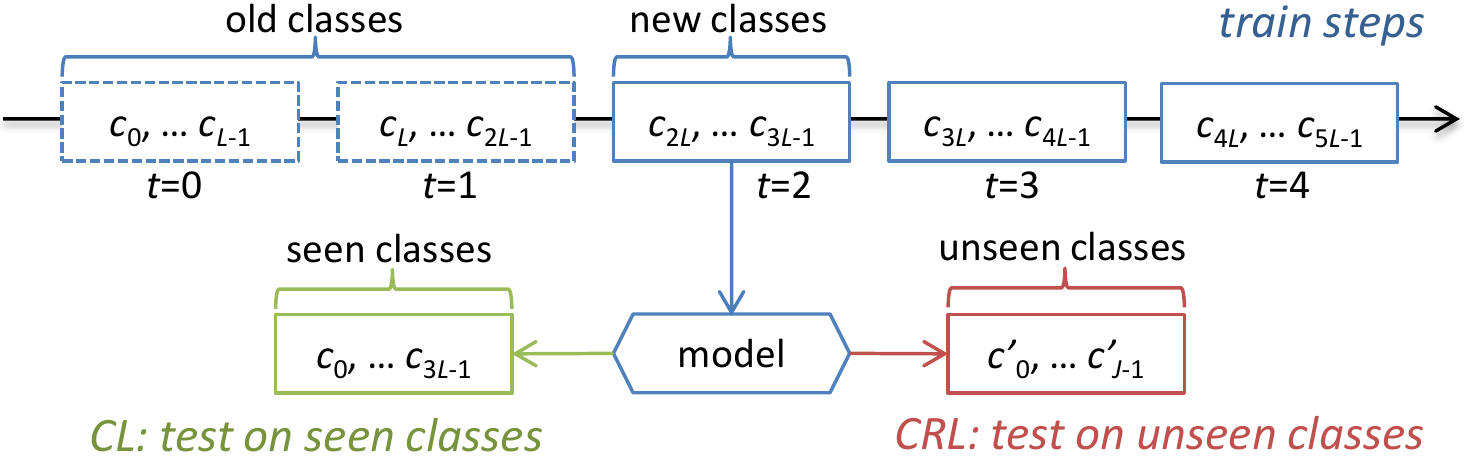}
\caption{\footnotesize{The proposed continual representation learning (CRL) v.s. the traditional continual learning (CL). The model is trained online on newly obtained tasks without access to old classes. Our CRL aims to learn better representation that is generalized to unseen classes, while traditional CL aims to learn and remember more old classes.}}
\label{fig:newsetting}
\end{figure}

In the standard offline learning paradigm, the model cannot preserve the previous knowledge well when being continuously trained on new tasks without access to old tasks, which is known as the Catastrophic Forgetting \cite{mccloskey1989catastrophic, ratcliff1990connectionist, mcclelland1995there, french1999catastrophic, goodfellow2013empirical}.
Continual learning (CL) \cite{shin2017continual, lopez2017gradient, rebuffi2017icarl, shmelkov2017incremental, castro2018end, parisi2019continual}  becomes an important research topic to alleviate such problem. 
For image recognition, continual learning is typically formulated as the class-incremental classification task.
The training process includes a sequence of training steps, and each step involves training with the images from new classes. 
Once the model is trained on the data of new classes, its performance is measured on a set of images from both old and new classes. The classes in the testing set are all previously seen (appeared) in the training set, thus the main goal is to recognize as many classes as possible without forgetting old classes.

Yet, this setting is not ideal for the biometric identification problems for various reasons.
First, biometric identification typically consists of train and test sets which are disjoint in terms of classes (or identities). The performance on learned old classes in CRL is easily kept high while learning new classes, which means CRL suffers little from the traditional forgetting problem, \emph{i.e.} forgetting old classes. Hence, it is not suitable to measure model's generalization ability on seen (old and new) classes in the new setting. We give the experimental evidence in Sec. \ref{sec:pre_exp}.
Thus, the main goal of our CRL is to generalize to previously unseen classes which is in contrast to the typical CL setting. The comparison of two settings is illustrated in Figure \ref{fig:newsetting}.
Second, biometric identification focuses on a more challenging setting which is similar to fine-grained classification \cite{yao2012codebook, wang2014learning, akata2015evaluation, xiao2015application}.  
The intra-class appearance variations are significantly subtler than the standard object classes in the commonly used CIFAR-100 \cite{krizhevsky2009learning} and ImageNet \cite{russakovsky2015imagenet} datasets. 
Hence, it is particularly challenging for continual representation learning, as the model has to learn better representation during many learning steps and improve the ability to discriminate unseen classes/identities.

Most existing CL benchmarks, illustrated in Table \ref{tab:cmp_benchmark}, are for either small-scale (e.g. MNIST \cite{lecun1998gradient}, CIFAR-100 \cite{krizhevsky2009learning}, CUB \cite{WahCUB_200_2011}) or coarse-grained (e.g. CORe50 \cite{lomonaco2017core50} and ImageNet \cite{russakovsky2015imagenet}) object recognition. 
To the best of our knowledge, there is no large-scale continual learning benchmark for biometric identification. To simulate the continuous stream setting for biometric identification,  we propose two large-scale benchmarks for face recognition and person re-id, which contain around 92K and 7K identities respectively. 
As shown in Table \ref{tab:cmp_benchmark}, the proposed two benchmarks are larger than all existing CL benchmarks when both class and image numbers are considered.
The identities of the training images are partitioned randomly and equally into 5/10 subsets, and each subset is used for one learning step. 
The identities of testing images are disjoint from those of training images, thus they can be fixed for all the steps and used to evaluate the generalization ability of learned representation.

The traditional continual learning methods usually learn classifiers for small-scale seen classes, and they are hardly scalable to a large number of identities in the real scenario. For example, the popular LWF method \cite{li2018learning} regularizes the consistency of outputs of all old classifiers in old and new models for knowledge distillation, in addition to minimizing the classification loss for learning new classes. A large number of identities will prohibit the usage of previous methods, because the limited memory and computation resources of GPU cannot handle the huge fully connected classification layer. To solve this problem, we also propose a method that implements knowledge distillation regarding the outputs of selected classes instead of all classes. In particular, the knowledge distillation is based on KL divergence instead of cross-entropy to regularize the difference between the outputs of old and new models. We then relax the regularization by an adaptive margin to give the model more flexibility to learn new knowledge. 

In summary, our contributions are two-fold: (1) We propose a new continual learning setting for learning better representation in biometric identification. Such setting requires a large-scale multi-step training set and a third-party testing set with identities that have never appeared in the training set. For this reason, we introduce two large-scale benchmarks for continual face recognition and continual person re-id. (2) To address the new setting, we propose a novel method with neighbourhood selection (NS) and consistency relaxation (CR) for knowledge distillation, which significantly improves the scalability and learning flexibility. Extensive experiments show that the representation can actually be improved in the continual representation learning setting by existing knowledge distillation strategies, and the proposed method achieves better results.

\begin{table*}
\centering
\scriptsize
\begin{tabular}{c|c|c|c|c|c|c|c|c|c}
\hline
\multirow{2}{*}{} & \multirow{2}{*}{Task}           & \multirow{2}{*}{Scale} & \multirow{2}{*}{Concept} & \multicolumn{3}{c|}{Classes} & \multicolumn{3}{c}{Images}      \\ \cline{5-10} 
                  &                                 &                        &                          & Train   & Test   & Total  & Train  & Test & Total      \\ \hline
MNIST \cite{lecun1998gradient}             & \multirow{5}{*}{Cls} & \multirow{4}{*}{Small} & number                   & 10         & 10        & 10     & 60,000    & 10,000  & 70,000     \\ \cline{1-1} \cline{4-10} 
CORe50 \cite{lomonaco2017core50}            &                                 &                        & coarse-grained objects   & 50         & 50        & 50     & 120,000   & 45,000‬ & 165,000    \\ \cline{1-1} \cline{4-10} 
CIFAR-100 \cite{krizhevsky2009learning}        &                                 &                        & coarse-grained objects   & 100        & 100       & 100    & 50,000    & 10,000  & 60,000     \\ \cline{1-1} \cline{4-10} 
CUB \cite{WahCUB_200_2011}              &                                 &                        & fine-grained birds       & 200        & 200       & 200    & 5,994     & 5,794   & ‭11,788‬   \\ \cline{1-1} \cline{3-10} 
ImageNet \cite{russakovsky2015imagenet}          &                                 & \multirow{3}{*}{Large} & coarse-grained objects   & 1,000      & 1,000     & 1,000  & 1,281,167 & 50,000  & 1,331,167‬ \\ \cline{1-2} \cline{4-10} 
\textit{CRL-face}     & \multirow{2}{*}{Rpt} &                        & fine-grained face        & 85,738     & 5,829        & 91,567       & 5,783,772          & 4,000        & 5,787,772           \\ \cline{1-1} \cline{4-10} 
\textit{CRL-person}     &                                 &                        & fine-grained person      & 2,494      & 4,512     & 7,006  & 59,706    & 30,927 & 90,633  \\ \hline
\end{tabular}
\caption{\footnotesize{Statistics of popular CL benchmarks and the proposed \textit{CRL-face} and \textit{CRL-person}. Cls: Classification. Rpt: Representation.}}
\label{tab:cmp_benchmark}
\end{table*}

\section{Related Work}
\label{sec:relatedwork}
\noindent \textbf{Biometric Identification.}
Much progress has been achieved in biometric identification including face recognition \cite{liu2015deep, wen2016discriminative, deng2019arcface} and person re-id \cite{li2014deepreid, xu2018attention, wei2018person}. Different from the coarse-grained object recognition, biometric identification involves much more fine-grained classes/identities. The recent improvements in biometric identification are achieved by learning better representation with different losses, \emph{e.g.}, softmax-based losses \cite{wen2016discriminative, deng2019arcface, li2014deepreid}, triplet-based losses \cite{schroff2015facenet, cheng2016person, hermans2017defense} and other kinds of losses \cite{varior2016gated, chen2017beyond}, on large-scale image datasets \cite{LFWTech, klare2015pushing, zheng2015scalable, guo2016ms, kemelmacher2016megaface, zheng2017unlabeled, wei2018person}. 

However, few works concern how to learn better representation from biometric data stream. The existing related works are different from our setting in terms of goal, training/testing protocol and dataset scale. 
\cite{ozawa2005incremental} proposed a method for incrementally updating the face recognition system by collecting mis-classified face images for further training. The method was evaluated on only 7 identities. 
Some methods \cite{sun2014online, wang2016human, martinel2016temporal} were proposed for online person re-id. Unfortunately, all observed training data need to be stored. In contrast, data of old classes are not accessible in our CRL setting. \cite{li2019one} proposed an online-learning method for one-pass person re-id. The feature extractor is pre-trained offline, and only the discriminator is updated online by new data. Besides, the model is not evaluated during every learning step. Overall, the tasks that these methods try to deal with are quite different from both the popular CL setting and our CRL setting.

\noindent \textbf{Continual Learning.}
Continual learning is also named lifelong learning \cite{silver2013lifelong, pentina2014pac, sodhani2018training}, incremental learning \cite{rebuffi2017icarl, shmelkov2017incremental, castro2018end} and sequential learning \cite{de2016incremental, aljundi2018selfless} in previous works. 
Existing continual learning works focus on general object recognition \cite{rebuffi2017icarl, kirkpatrick2017overcoming, castro2018end}, object detection \cite{shmelkov2017incremental, guan2018learn}), image generation \cite{wu2018memory, lesort2018generative}, reinforcement learning tasks \cite{kaplanis2018continual, abel2018state, xu2018reinforced} and unsupervised learning tasks \cite{Rao2019unsupervised}. 
The popular continual learning setting is to continuously learn new classes and test on all seen (both old and new) classes, and it suffers from the catastrophic forgetting problem.
\cite{aljundi2019task} proposed ``Task-free continual learning'' in which data classes in different learning steps may be joint. Like the popular setting, they train and test on the same classes, and the number of classes is quite limited.

A number of methods are proposed to avoid the catastrophic forgetting of deep models. Generally speaking, they can be divided into two kinds. The one is based on rehearsal \cite{rebuffi2017icarl, lopez2017gradient, castro2018end} or pseudo-rehearsal \cite{shin2017continual, wu2018memory, kangcontinual}, which requires an extra memory or generative model to remember old task data. The other one is based on the regularization on weights \cite{kirkpatrick2017overcoming, zenke2017continual, aljundi2018memory}, features \cite{jung2016less}, and outputs \cite{li2018learning, castro2018end}.


The popular benchmarks for evaluating these CL methods are (original or permuted) MNIST \cite{lecun1998gradient}, CORe50 \cite{lomonaco2017core50}, CIFAR-100 \cite{krizhevsky2009learning}, CUB \cite{WahCUB_200_2011} and ImageNet \cite{russakovsky2015imagenet}. 
Except ImageNet (with 1K classes and 1.3M images), all other benchmarks are small-scale in terms of class ($\leq 200$) and image ($\leq 170$K) numbers. These benchmarks cannot reflect the necessity of implementing continual learning, because they can be easily handled by limited computing resources.
In addition, except CUB, all benchmarks are about coarse-grained objects. Hence, they are not suitable for evaluating the representation ability of deep models. 

Different from the popular CL setting, the proposed CRL aims to continuously learn more generalized representation model for identifying many unseen classes/identities. The proposed two benchmarks are the first large-scale benchmarks for CRL, and they are much larger than existing CL benchmarks in terms of class (92K and 7K) and image (5.8M and 91K) numbers.

\section{CRL Setting and Benchmarks}

\textbf{CRL Setting.} As illustrated in Figure \ref{fig:newsetting}, the model will be trained for in total $T=5$ steps starting from step $0$. Each training step includes $L$ classes, and training classes of different steps are disjoint. The model can only access the training data of current learning step $t$. For example, assuming current step $t=2$, the model is trained only on training data of new classes $c_{2L}, \dots c_{3L-1}$. Without accessing old classes $c_{0}, \dots c_{2L-1}$, the model will gradually forget the knowledge obtained from previous learning steps. In each step, CRL tests the model on previously unseen testing classes $c'_{0}, \dots c'_{J-1}$ for evaluating model's generalization ability, which is frequently used as the performance metric in biometric identification tasks.

We present continual representation learning benchmarks for two popular biometric identification tasks, namely, face recognition and person re-id. The statistics of the two benchmarks are shown in Table \ref{tab:cmp_benchmark}. The data can be found in GitHub Project\footnote{https://github.com/PatrickZH/Continual-Representation-Learning-for-Biometric-Identification}. The presented benchmarks are different from existing continual learning benchmarks in three main aspects.
\begin{itemize}
\item The proposed CRL benchmarks are specifically designed for biometric (face or person) identification, while existing CL benchmarks focus on the general image classification.
\item The number of classes in our benchmarks (92K and 7K) is much larger than existing benchmarks ($\leq$ 1K).
\item We test the model on novel identities that have never appeared in the training set, while existing benchmarks test on new images of learned (seen) classes.
\end{itemize}

\noindent \textbf{CRL-face Benchmark} Continual face recognition requires large-scale training data for each learning step. Ms1M dataset \cite{guo2016ms} is a suitable option for constructing CRL-face benchmark, because there are 85,738 identities and 5,783,772 images in the dataset. We divide identifies in Ms1M into 5 and 10 subsets randomly and equally. Each split subset has around 17,148 identities for 5-step setting and 8,573 identities for 10-step setting, respectively. The number of images in each subset varies because the number of images associated with each identity is not equal. Each subset serves as the training set in each learning step.

Two testing datasets, namely LFW \cite{LFWTech} and Megaface \cite{kemelmacher2016megaface}, are used for evaluating the representation ability of models. The LFW dataset is the most widely used as the testing benchmark that contains 6,000 testing pairs from 5,749 identities. We follow the unrestricted with labelled outside data protocol, where features are trained with additional data and the verification accuracy is estimated by a 10-fold cross validation scheme. 9 folders are combined as the validation set to determine the threshold , and the 10th folder is used for testing. The Megaface benchmark is another challenge for face recognition. It contains 1M images of 690K different individuals as the gallery set and 100K photos of 530 unique individuals from FaceScrub as the probe set. For testing, the target set has 4000 images of 80 identities, and the distractor set has over 1M images of different identities. The Top 1 accuracy is reported. 


\noindent \textbf{CRL-person Benchmark} To obtain enough identities for implementing continual learning, we combine three popular person re-id datasets, namely, Market1501 \cite{zheng2015scalable}, DukeMTMC-reID  \cite{zheng2017unlabeled} and MSMT17\_V2 \cite{wei2018person}. The mixed dataset (CRL-person) contains 2,494 training identities. Specifically, the three datasets contribute 751, 702 and 1,041 training identities respectively. In total, 59,706 training images of the 2,494 identities are employed as the training set. The training set is split into 5 subsets and 10 subsets for 5-step and 10-step continual representation learning respectively.

For testing, we combine the testing sets of the three datasets. However, evaluating the model on all testing data is computational expensive, as there are 17,255 query images and 119,554 gallery images in total. Thus, we apply two strategies to reduce the image number: (1) We keep all query identities and remove those identities only in the gallery set. (2) We randomly select at most one image for each identity under each camera in query and gallery set respectively. After applying the two strategies, the final testing set has 11,351 query images and 19,576 gallery images of 4,512 identities. Mean average precision (mAP) and top1 accuracy are reported in each learning step.  

\section{Flexible Knowledge Distillation for CRL}
According to our protocol, we need to continuously train our identification model for multi-steps. The $t$th step provides new data $\mathcal{O}^{t}=\{(x^t_{i}, y^t_{i})  \}_{i=1}^{n}$, where each instance $(x^t_{i}, y^t_{i})$ is composed by an image $x^t_{i}\!\in\!\mathcal{X}^{t}$ and a label $y^t_{i}\!\in\!\mathcal{Y}^{t}$. $n=|\mathcal{O}^{t}|$ is the number of all new data. The goal of 
CRL is to construct an embedding function $f$, which can compute a feature representation $\phi_{i}$ to better associate with $y^t_{i}$.  To accomplish this, we consider the  $f(x^t_{i}, \theta^{t})$ parameterized by $\theta^{t}$, and define a classification loss based on the $t$th step data $\mathcal{O}^{t}$:
\begin{equation}\small
\begin{split}
&\mathcal{L}^{new}( \theta^{t}, \mathbf{w}^{t}; \mathcal{O}^{t})  = \frac{1}{n} \sum_{i=1}^{n} l(\phi_{i}, y^t_{i}, \theta^{t}, \mathbf{w}^{t} ), \\
&\phi_{i} =  f(x^t_{i}, \theta^{t}),\\
&l(\phi_{i}, y^t_{i}, \theta^{t}, \mathbf{w}^{t}) \!=\! - \log  \frac{\exp( \phi_{i}^{\top} \mathbf{w}^{t}_{y^t_{i}})}{\sum_{j=1} \exp(\phi_{i}^{\top} \mathbf{w}^{t}_{j} )} .
\end{split}\label{eq:obj1}
\end{equation}
$w^t_j$ indicates the classifier for the $j$th class.
Obviously, minimizing the loss in Eq. \ref{eq:obj1} will result in overfitting to the instances in $\mathcal{O}^{t}$. As an alternative, we could additionally maintain a memory data set to keep the predictions at the past steps invariant, which will lead to the problem on how to select the most useful samples from the past data.  This paper focuses on the scenarios where there are no memory data. We only have the model $f(x^t_{i}, \theta^{t-1})$ and classifiers $\mathbf{w}^{t-1}$ in the last step. It is suitable to employ knowledge distillation(KD) to optimize a loss function based on the old model and the current data. 

\begin{figure*}[t!]
\centering
\includegraphics[width=0.85\linewidth]{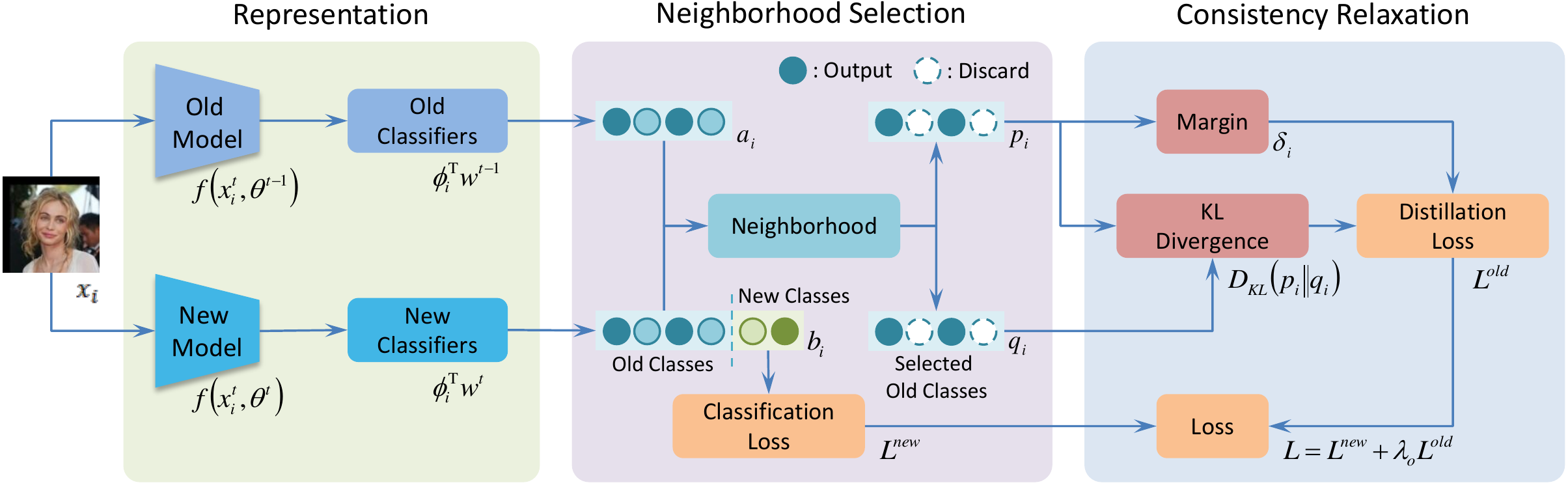}
\caption{\footnotesize{Illustration of the proposed method. Our method consists of three modules, namely representation, neighborhood selection and consistency relaxation. 
Representation: The image $x^t_i$ is fed into both old and new models, $f(x^t_{i}, \theta^{t-1})$ and $f(x^t_{i}, \theta^{t})$, followed by corresponding old and new classifiers $\mathbf{w}^{t-1}$ and $\mathbf{w}^{t}$. The activation of old and new models, $\mathbf{a}_{i}$ and $\mathbf{b}_{i}$, are produced. We use the activation of new classes produced by the new model to calculate the classification loss $\mathcal{L}^{new}$.
Neighborhood selection: We determine the neighborhood of the given datum based on activation of old model $\mathbf{a}_{i}$ and choose top ones of old and new models in the neighborhood to calculate KL divergence $\mathcal{D}_{KL}(\mathbf{p}_{i} || \mathbf{q}_{i})$. 
Consistency Relaxation: The margin $\delta_{i}$ is introduced to KL divergence for consistency relaxation, and the relaxed KL divergence is produced as the distillation loss $\mathcal{L}^{old}$. The overall loss $\mathcal{L}$ is the weighted combination of the classification loss $\mathcal{L}^{new}$ and distillation loss $\mathcal{L}^{old}$.}}
\label{fig:method}
\end{figure*}

\subsection{Knowledge Distillation}
The idea of knowledge distillation was found by Hinton \emph{et al}, which works well for encouraging the output of one network to approximate that of another network. Suppose $\mathbf{w}^{t-1}$ is about $L$ classes, the output probability of $y^t_{i}$ generated by the old model $f(x^t_{i}, \theta^{t-1})$ given $\mathbf{w}^{t-1}$ is: $\mathbf{p}_{i} = \{ p_{i,1}, p_{i,2},...,p_{i,L}\}$. The cross-entropy loss is utilized to regularize the new probability $\mathbf{q}_{i} = \{q_{i,1}, q_{i,2},...,q_{i,L} \}$ generated by the new model $f(x^t_{i}, \theta^{t})$: 
\begin{equation}
 \mathcal{L}^{old}(\theta^{t},\mathbf{w}^{p}; \mathcal{O}^{t}) = -\frac{1}{n} \sum_{i=1}^{n} \sum_{l=1}^{L} p_{i,l} \log q_{i,l}, \label{Eq:loss_old}
\end{equation}
where $l$ is the index of the class. 

\noindent  \textbf{Discussion}. Compared with the existing CL scenarios, a crucial difference for the proposed setting is that it handles a large number of classes/identities. For example, existing CL methods are evaluated  on 10 classes of MNIST, 100 classes of CIFAR and 1,000 classes of ImageNet, while the model for biometric identification usually needs to be trained over thousands to millions classes. The scalability and efficiency of training methods become important due to the limited memory and computation resources. Furthermore, the current knowledge distillation require the strict consistency between the outputs of new and old models, and it largely restricts the ability to learn new knowledge. Based on the above concerns, we propose Flexiable Knowledge Distillation(FKD), where we perform  neighbour selection and consistency relaxation over the loss related to the old model $\mathcal{L}^{old}$. The proposed method is illustrated in Figure \ref{fig:method}.

\subsection{Neighbourhood Selection}
 In the standard  knowledge distillation(Eq.\ref{Eq:loss_old}),  the  probability distribution is calculated based on the activation  (the direct output of the classifiers) by a softmax layer. When the number of classifies increases to thousands even millions, its not scalable for maintaining such a large fully-connected classifier layer. At the same time, the computation of softmax probability is not effective for such a large number of classes,  as the probability values are weakened by many unrelated classes.  Hence,  we select a few ``similar" classes from all old classes to implement selected knowledge distillation.   
 
 Given a sample $x^t_{i}$, the activation generated by the old model is denoted by $\mathbf{a}_{i} = \{a_{i,1}, a_{i,2}, ..., a_{i,L} \}$ and the activation of the new model is denoted by $\mathbf{b}_{i} = \{ b_{i,1}, b_{i,2}, ..., b_{i,L} \}$.  We rank the activation units of the old model ($\mathbf{a}_{i}$) with descending order, select the top $K$ ones, and put their indices into the set $\mathcal{S}_{i}$, \emph{i.e.}, the neighborhood of the ground-truth class $y^t_i$. The probabilities generated by the old and new model are $\mathbf{p}_{i}$ and $\mathbf{q}_{i}$, which are calculated based on the selected label set $\mathcal{S}_{i}$:
 \begin{equation}
   \begin{split}
    p_{i,l} = \frac{\exp(a_{i,l}/T)}{ \sum_{j \in \mathcal{S}_{i}} \exp(a_{i,j}/T) }, \;\;\;\;\;
    q_{i,l} = \frac{\exp(b_{i,l}/T)}{ \sum_{j \in \mathcal{S}_{i}} \exp(b_{i,j}/T)}, 
   \end{split}
   \label{eq:probability}
 \end{equation}
 where $T$ is the hyper-parameter of knowledge distillation. Instead of using cross-entropy loss, we utilize the Kullback Leibler(KL) divergence, \emph{i.e.} KLD, to measure the difference between $\mathbf{p}_{i}$ and $\mathbf{q}_{i}$:
 \begin{equation}
     \mathcal{D}_{KL}(\mathbf{p}_{i} || \mathbf{q}_{i}) = \sum_{l \in \mathcal{S}_{i} } ( p_{i,l} \log p_{i,l} - p_{i,l} \log q_{i,l} ). 
\label{eq:kld}
\end{equation}
As $ \sum_{l \in \mathcal{S}_{i} } p_{i,l} \log p_{i,l}$ is a constant in the optimization, the KL-divergence is equivalent to the cross-entropy in Eq. \ref{Eq:loss_old}, and $\mathcal{D}_{KL}(\mathbf{p}_{i} || \mathbf{q}_{i})$ will be 0 if $\mathbf{p}_{i}$ and $ \mathbf{q}_{i}$ are the same.

\subsection{Consistency Relaxation}
The new model needs to learn knowledge from both old and new classes. The best parameters of new model should not be exactly the same as the old model. Hence, we introduce an adaptive margin $\delta_{i}$ to relax the consistency constraint. The margin for $\mathbf{x}_{i}$ is set to be:
\begin{equation}
    \delta_{i} = -\beta \sum_{l \in \mathcal{S}_{i}} p_{i,l} \log p_{i,l}, 
\label{eq:margin}
\end{equation}
where $\beta$ is the coefficient that controls the  magnitude  of margin and the term $- \sum_{l \in \mathcal{S}_{i}} p_{i,l} \log p_{i,l}$ is the minimal value of cross-entropy $- \sum_{l \in \mathcal{S}_{i}} p_{i,l} \log q_{i,l}$. With the margin, the KL-divergence is relaxed by:
\begin{equation}
\mathcal{D}'_{KL}(\mathbf{p}_{i}|| \mathbf{q}_{i}) =  [\mathcal{D}_{KL}(\mathbf{p}_{i}|| \mathbf{q}_{i}) - \delta_{i}]_{+}, 
\label{eq:relaxed_KLD}
\end{equation}
where $[\cdot]_{+}$ indicates the hinge loss. Minimizing the relaxed KL-divergence \\ $\mathcal{D}'_{KL}(\mathbf{p}_{i}|| \mathbf{q}_{i})$ indicates the cross-entropy should be as small as $-\sum_{l \in \mathcal{S}_{i}} p_{i,l} \log p_{i,l} $ until it is smaller than $-(1+\beta) \sum_{l \in \mathcal{S}_{i}} p_{i,l} \log p_{i,l}$. With the selection and relaxation, the loss term $\mathcal{L}^{old}(\theta^{t},\mathbf{w}^{p}; \mathcal{O}^{t})$ can be reformulated by:
\begin{equation}
   \mathcal{L}^{old}(\theta^{t},\mathbf{w}^{t}; \mathcal{O}^{t}) = -\frac{1}{n} \sum_{i=1}^{n} \mathcal{D}'_{KL}(\mathbf{p}_{i}|| \mathbf{q}_{i}) \label{eq:obj2}
\end{equation}

\subsection{Learning Algorithm}
The overall objective function combines the classification loss (Eq.\ref{eq:obj1}) and flexible knowledge distillation loss (Eq. \ref{eq:obj2}) by a  balance weight $\lambda_{0}$, which is used to optimize $\theta^{t}, \mathbf{w}^{t}, \mathbf{w}^{p}$:
\begin{equation}
\begin{split}
   \mathcal{L} =  \mathcal{L}^{new}(\theta^{t}, \mathbf{w}^{t}; \mathcal{O}^{t}) + \lambda_{o}  \mathcal{L}^{old}(\theta^{t},\mathbf{w}^{t}; \mathcal{O}^{t}).
\end{split}
\label{eq:final}
\end{equation}

Algorithm \ref{alg} shows the main steps for training the new model and classifiers. First, we initialize the new model $\theta^{t}$ and classifiers $w^{t}$ by copying weights from the old model $\theta^{t-1}$ and classifiers $w^{t-1}$. As the number of new classifiers increases, we randomly initialize the added weights. For a batch of taining data, we compute the classification loss $\mathcal{L}^{new}$ using Eq. \ref{eq:obj1}. Then, for each datum $(x^t_i, y^t_i)$, we do the flexible knowledge distillation. The activation $\mathbf{a}_i=\{a_{i,1},a_{i,2},..,a_{i,L}\}$ and $\mathbf{b}_i=\{b_{i,1},b_{i,2},..,b_{i,L}\}$ are produced by the old and new models. The valid units in the neighborhood $\mathcal{S}_{i}$ are selected as the top $K$ ones in $\mathbf{a}_i$, and the rest units are ignored. We also select the corresponding units for $\mathbf{b}_i$ based on $\mathcal{S}_{i}$. With the selected units, we compute the probabilities $\mathbf{p}_{i}$ and $\mathbf{q}_{i}$ using Eq. \ref{eq:probability}. Then,
KL divergence $\mathcal{D}_{KL}$ and margin $\delta_{i}$ are calculated using Eq.  \ref{eq:kld} and \ref{eq:margin} for obtaining the relaxed KL divergence $\mathcal{D}'_{KL}$ using Eq. \ref{eq:relaxed_KLD}. The distillation loss $\mathcal{L}^{old}$ for the batch is calculated by Eq. \ref{eq:obj2}. The final loss $\mathcal{L}$ is calculated as the weighted combination of $\mathcal{L}^{new}$ and $\mathcal{L}^{old}$ using Eq. \ref{eq:final}. With loss $\mathcal{L}$, we update $\theta^{t}$ and $w^{t}$ by back-propagation.

\begin{algorithm}
\footnotesize
\caption{\footnotesize{Flexible Knowledge Distillation}\label{alg:euclid}}
\begin{algorithmic}[1]
\Require 
    \Statex $\mathcal{O}^{t}=\{(x^t_{i}, y^t_{i})  \}_{i=1}^{n}$ : training data in current learning step; \Statex $\theta^{t-1}$ : old model; $w^{t-1}$ : old classifiers;
\Ensure 
    \Statex $\theta^{t}$ : new model; $w^{t}$ : new classifiers;
\State Initialize $\theta^{t}$, $w^{t}$ by $\theta^{t-1}$, $w^{t-1}$;
\For{a batch in $\mathcal{O}^{t}$}
    \State Compute classification loss $\mathcal{L}^{new}$ using Eq. \ref{eq:obj1};
    \For{$(x^t_i, y^t_i)$ in the batch}
        \State Determinate the neighborhood $\mathcal{S}_i$ based on activations $\mathbf{a}_i$ (old) and $\mathbf{b}_i$ (new);
        \State Compute relaxed KL Divergence $\mathcal{D}'_{KL}$ using Eq. \ref{eq:relaxed_KLD};
    \EndFor
    \State Compute distillation loss $\mathcal{L}^{old}$ using Eq. \ref{eq:obj2} and the final loss $\mathcal{L}$ using Eq. \ref{eq:final};
    \State Update $\theta^{t}$ and $w^{t}$ by back-propagation with $\mathcal{L}$.
\EndFor
\end{algorithmic}
\label{alg}
\end{algorithm}

\section{Experiments}
\label{sec:experiments}
In this section, we present implementation details and results on the two CL benchmarks, namely, CRL-face and CRL-person. A preliminary experiment is given for illustrating why performance on old classes is not suitable to measure CRL methods. In addition, the ablation study is given, which verifies the importance of two modules in our method.

\subsection{Implementation Details}
We use ResNet-50 as the backbone for both continual face recognition and person re-id. The temperature $T$ is set to be $2$ for all knowledge distillation based methods in experiments. The balance weight $\lambda_o$ is chosen from $10^{\{-2,-1,0,1\}}$ for different methods individually. We use hold-off validation data to determine the two hyper-parameters ($K$ and $\beta$), we first select the best $K$ without CR module, then we choose the best $\beta$ based on the selected $K$. The details will be given in Sec. \ref{sec:ablation}. The classification/retrieval of testing data is based on the similarity (Euclidean distance) of feature embeddings. For all experiments, we repeat five times and report the mean value and standard deviation.

\noindent \textbf{Continual Face Recognition.} All images are aligned and then resized to $112 \times 112$. The feature dimension is 256, and the batch size is 384. SGD optimizer with initial learning rate $10^{-2}$ is used in face experiments. For 5-step setting, we train the model for 20000 iterations in each learning step, and the learning rate is reduced by $\times 0.1$ at $8000$th and $16000$th iteration. For 10-step setting, the model is trained for 10000 iterations and the learning rate is reduced by $\times 0.1$ at $4000$th and $8000$th iteration. 

\noindent \textbf{Continual Person Re-id.} All images are resized to $256\times 128$. The feature dimension is 2,048, and the batch size is $256$. Following the popular person re-id training strategy, in each training batch, we randomly select $64$ identities and sample 4 images for each identity. Adam optimizer with learning rate $3.5\times 10^{-4}$ is used. We train the model for 50 epochs, and we decrease the learning rate by $\times 0.1$ at the $25$th and $35$th epoch.

\subsection{Preliminary Experiment}
\label{sec:pre_exp}
We first give a simple preliminary experiment to illustrate that Catastrophic Forgetting of old classes is not the main problem of CRL and the performance on old classes is not suitable to evaluate CRL methods. In this experiment, the model is continually trained (or finetuned) on 5 subsets of CRL-face and evaluated on the hold-off testing data of the first subset (Step0). In other words, the model is always evaluated on classes of Step0.
According to Table \ref{tab:pre_exp}, the performance on old classes of the first subset does not show obvious decrease even simply finetuning the model on new classes without any regularization. The performance even increased on Step1 due to learning new classes. It means that CRL models suffer little from Catastrophic Forgetting of old classes. Hence, we evaluate model's generalization ability on unseen classes in CRL setting, which is more suitable. 

\begin{table}
\centering
\footnotesize
\begin{tabular}{cccccc}
\hline
         & Step0 & Step1 & Step2 & Step3 & Step4 \\ \hline
Finetune & 93.38 & 95.05 & 94.20  & 94.08  & 93.82  \\ \hline
\end{tabular}
\caption{\footnotesize{The performance (\%) on old classes on CRL-face dataset. Note that the training set of CRL-face is split into 5 subsets with 17,148 classes per subset. The performances are evaluated on the classes of the first subset (Step0).}}
\label{tab:pre_exp}
\end{table}

\begin{figure*}[]
\centering
\includegraphics[width=0.95\linewidth]{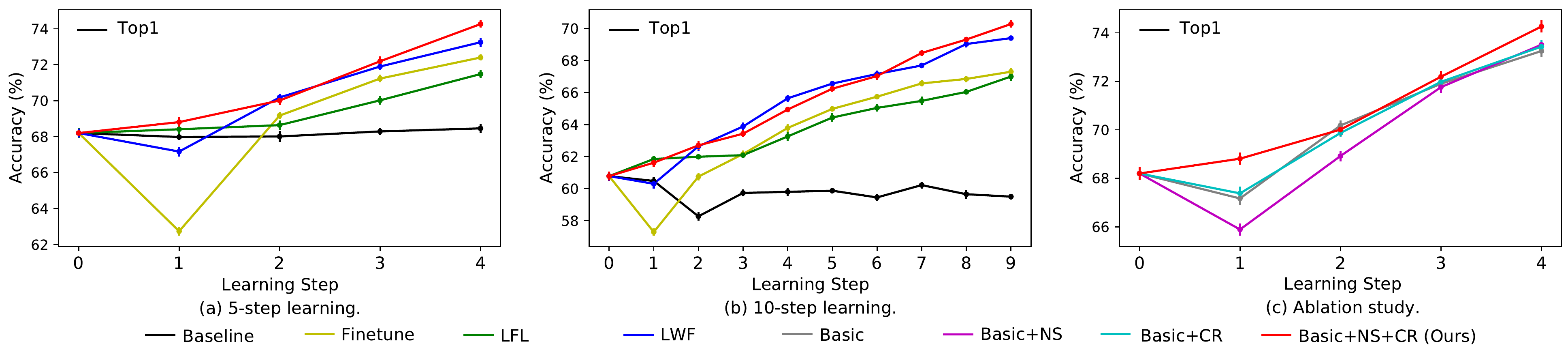}
\caption{\footnotesize{Experimental results on continual face recognition (tesed on Megaface). Top 1 accuracy (\%) is reported. Figure (a) and (b) are 5-step and 10-step continual learning. We compare our method to Baseline, Finetune, LFL and LWF. Figure (c) is the ablation study. We compare the variants of our method with/without Neighborhood Selection (NS) and Consistency Relaxation (CR) modules.}}
\label{fig:exp_face}
\end{figure*}

\begin{table*}
\centering
\scriptsize
\begin{tabular}{cccccccc}
\hline
                          &         & Baseline & Finetune & LFL   & LWF     & Ours    & Upper-bound             \\ \hline
\multirow{2}{*}{5-step} & LFW       & $98.85\pm0.02$ & $99.00\pm0.01$ & $98.97\pm0.02$ & $98.95\pm0.01$ & $\mathbf{99.10\pm0.01}$ & $99.42\pm0.01$ \\ \cline{2-8}
                        & Megaface  & $68.20\pm0.21$ & $72.40\pm0.12$ & $71.48\pm0.23$ & $73.25\pm0.22$ & $\mathbf{74.26\pm0.23}$ & $82.93\pm0.12$ \\ \hline
\multirow{2}{*}{10-step}& LFW       & $98.52\pm0.02$ & $98.72\pm0.04$ & $98.65\pm0.02$ & $98.82\pm0.02$ & $\mathbf{99.05\pm0.01}$ & $99.42\pm0.01$ \\ \cline{2-8}
                        & Megaface  & $60.78\pm0.22$ & $66.85\pm0.21$ & $67.00\pm0.12$ & $69.03\pm0.24$ & $\mathbf{70.28\pm0.13}$ & $82.93\pm0.12$ \\ \hline
\end{tabular}
\caption{\footnotesize{Comparison to SOTA: face recognition. Top 1 accuracy ($\%$) in the final learning step. Upper-bound means joint training.}}
\label{tab:exp_face}
\end{table*}

\subsection{Comparison to the State-of-the-art}
\begin{figure*}{}
\centering
\includegraphics[width=0.95\linewidth]{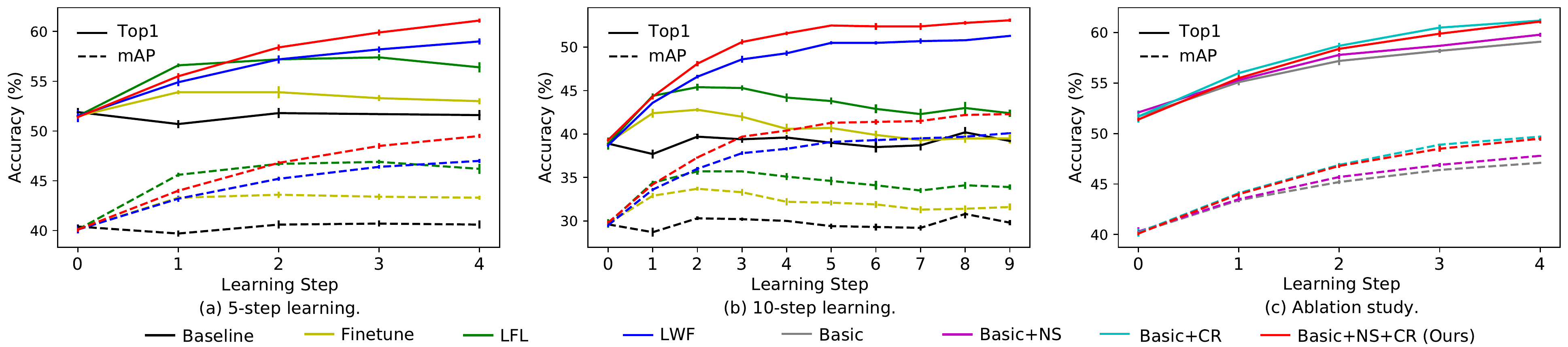}
\caption{\footnotesize{Experimental results on continual person re-id. Top 1 accuracy and mAP (\%) are reported. Figure (a) and (b) are 5-step and 10-step continual learning. We compare our method to Baseline, Finetune, LFL and LWF. Figure (c) is the ablation study. We compare the variants of our method with/without Neighborhood Selection (NS) and Consistency Relaxation (CR) modules.}}
\label{fig:exp_reid}
\end{figure*}

Our goal is not improving SOTA face recognition or person reid performance, instead we aims to extend continual learning to biometric identification and propose a scalable CL method. Hence, we compare to those SOTA CL methods with the same backbone and task (classification) loss.
We provide the comparison to Baseline, Finetune and two SOTA CL methods, namely, LFL \cite{jung2016less} and LWF \cite{li2018learning}. 
We choose LFL and LWF as competitors because they are efficient enough for large-scale training on 5.8M images and 86K classes, while those generative model \cite{shin2017continual, wu2018memory, xiang2019incremental}, meta-learning \cite{hu2018overcoming, vuorio2018meta, javed2019meta, he2019task} and dynamic-network \cite{yoon2017lifelong, munkhdalai2017meta, aljundi2018selfless} based methods are not salable or efficient to train on such large benchmarks.
Baseline means that the model is trained from scratch (without old model) in every learning step. In this method, the model totally forgets knowledge learned from old classes. Finetune is a naive continual learning method in which the model is updated by finetuning the old model on new classes. LFL aims to restrict the difference of features produced by old and new models. In this way, the new model can produce similar features like the old model. LWF is based on knowledge distillation, which minimizes the cross-entropy between the outputs of old and new models. We choose to compare to LWF, because it is a representation of knowledge distillation based methods, \emph{e.g.}, iCaRL \cite{rebuffi2017icarl} and End2End \cite{castro2018end}.
We also provide the \emph{upper-bound} of each experimental setting. The upper-bound is calculated by jointly training on all data (of all steps). The performance in the final learning step is given in Table \ref{tab:exp_face} and \ref{tab:exp_reid}, while the detailed results of every learning step are illustrated in Figure \ref{fig:exp_face} and \ref{fig:exp_reid}.

\noindent \textbf{Continual Face Recognition.} Table \ref{tab:exp_face} shows the final results of our method on LFW and Megaface, compared to the state-of-the-art. Clearly, on the same dataset, performance of 10-step learning is worse than that of 5-step learning, because of fewer training data per step. Generally speaking, our method outperforms all other methods in all settings. Especially, when tested on Megaface, ours overwhelms the runner-up (LWF) by 1.01\% and 1.25\% on 5-step and 10-step settings respectively. As 5-step learning on LFW dataset is easy, Finetune also achieves good performance. However, on harder setting (10-step learning) and dataset (Megaface), the gap between Finetune and others widens.
Figure \ref{fig:exp_face} (a) and (b) illustrate the performance evaluated on Megaface in every learning step. We find that, except Baseline, performance of all methods increases after learning more classes. Our method shows obvious advantages compared to others.


\begin{table*}
\centering
\scriptsize
\begin{tabular}{cccccccc}
\hline
                         &      & Baseline      & Finetune      & LFL           & LWF       & Ours          & Upper-bound \\ \hline
\multirow{2}{*}{5-step}  & Top1 & $51.6\pm0.5$ & $53.0\pm0.3$ & $56.4\pm0.5$ & $59.0\pm0.3$ & $\mathbf{61.1\pm0.2}$ & $75.5\pm0.1$ \\ \cline{2-8} 
                         & mAP  & $40.6\pm0.4$ & $43.3\pm0.2$ & $46.2\pm0.5$ & $47.0\pm0.2$ & $\mathbf{49.5\pm0.2}$ & $64.6\pm0.1$ \\ \hline
\multirow{2}{*}{10-step} & Top1 & $39.2\pm0.3$ & $39.5\pm0.6$ & $42.4\pm0.4$ & $51.3\pm0.1$ & $\mathbf{53.1\pm0.2}$ & $75.5\pm0.1$ \\ \cline{2-8} 
                         & mAP  & $29.8\pm0.3$ & $31.6\pm0.4$ & $33.9\pm0.3$ & $40.1\pm0.1$ & $\mathbf{42.3\pm0.2}$ & $64.6\pm0.1$ \\ \hline
\end{tabular}
\caption{\footnotesize{Comparison to SOTA: person re-id. Top 1 and mAP accuracy ($\%$) in the final learning step. Upper-bound means joint training.}}
\label{tab:exp_reid}
\end{table*}

\noindent \textbf{Continual Person Re-id.} Table \ref{tab:exp_reid} shows Top1 and mAP performance of 5-step and 10-step learning settings on the proposed dataset. The gap between different methods is obvious. Our method outperforms the runner-up (LWF) by around $2\%$ on all settings. Compared to Baseline, our method improves the performance by $13.9\%$ (Top1) and $12.5\%$ (mAP) on 10-step learning, which means our method effectively leverages knowledge from old classes. However, our results are still obviously lower than the upper-bound. The gap indicates the challenging of continual person re-id on the proposed benchmark. Figure \ref{fig:exp_reid} (a) and (b) illustrate the performance (Top1 and mAP) of different methods in every learning step. Our method shows obvious advantage compared to other methods, especially on the 10-step learning. It is interesting that, after several learning steps, Finetune and LFL cannot effectively improve the performance on third-party testing classes when learning new training classes. However, knowledge distillation based methods (LWF and ours) can continuously improve performance. One possible reason is that the restriction on features prevents the model from learning better representation of data. 


\subsection{Ablation Study}
\label{sec:ablation}
\noindent \textbf{Effectiveness of Proposed Modules.}
We do ablation study on two modules of the proposed method, namely, Neighborhood Selection (NS) and Consistency Relaxation (CR). To verify the effectiveness of two modules, we compare the four variants of our method: (1) \textbf{Basic}: Plain knowledge distillation without NS or CR; (2)\textbf{Basic+NS}: Basic with Neighborhood Selection; (3) \textbf{Basic+CR}: Basic with Consistency Relaxation; (4) \textbf{Basic+NS+CR} (Ours): Basic with both Neighborhood Selection and Consistency Relaxation.

We do ablation study on 5-step continual face recognition and person re-id. Table \ref{tab:exp_abl_face} and \ref{tab:exp_abl_reid} show the results on the two benchmarks. Clearly, both NS and CR modules benefit the final performance. 
The improvement is obvious in continual person re-id. By adding NS module, Basic+NS overwhelms Basic by $0.7\%$ of Top1 and $0.7\%$ of mAP. Meantime, Basic+CR outperforms Basic by $2.1\%$ of Top1 and $2.6\%$ of mAP. 
In continual face recognition, Basic+NS+CR outperforms Basic+NS by $0.13\%$ and $0.83\%$ on LFW and Megaface respectively.

\begin{table}
\centering
\scriptsize
\setlength{\tabcolsep}{2pt}
\begin{tabular}{ccccc}
\hline
 & Basic    & Basic+NS & Basic+CR & \begin{tabular}[c]{@{}c@{}}Basic+NS+CR (Ours)\end{tabular} \\ \hline
LFW         & $98.95\pm0.02$ & $99.05\pm0.02$ & $98.97\pm0.01$ & $99.10\pm0.01$ \\ \hline
Megaface    & $73.25\pm0.31$ & $73.51\pm0.22$ & $73.43\pm0.24$ & $74.26\pm0.22$ \\ \hline
\end{tabular}
\caption{\footnotesize{Ablation study: face recognition. The Top 1 accuracy of different variants of our method, \emph{i.e.}, with/without NS and CR modules.}}
\label{tab:exp_abl_face}
\end{table}

\begin{table}
\centering
\scriptsize
\begin{tabular}{ccccc}
\hline
& Basic    & Basic+NS & Basic+CR & \begin{tabular}[c]{@{}c@{}}Basic+NS+CR (Ours)\end{tabular} \\ \hline
Top1 & $59.1\pm0.1$ & $59.8\pm0.2$ & $61.2\pm0.2$ & $61.1\pm0.2$                                                     \\ \hline
mAP  & $47.1\pm0.1$ & $47.8\pm0.1$ & $49.7\pm0.1$ & $49.5\pm0.2$                                                    \\ \hline
\end{tabular}
\caption{\footnotesize{Ablation study: person re-id. The Top 1 accuracy and mAP of different variants of our method, \emph{i.e.}, with/without NS and CR modules.}}
\label{tab:exp_abl_reid}
\end{table}

Figure \ref{fig:exp_face} (c) shows the results of four variants in continual face recognition. NS module may hinder knowledge transfer in first three steps because of fewer distillation bases (selected old classes). Finally, Basic+NS outperforms Basic. Basic+NS+CR has the best performance compared to other variants.
Figure \ref{fig:exp_reid} (c) illustrates how the two modules influence continual person re-id performance. We find that NS module of Basic+NS stably improves the performance compared to Basic. In contrast, some softmax selection methods \cite{grave2017efficient, blanc2017adaptive} in NLP are only for speeding up with the cost of performance decrease.
Besides, CR module also significantly and stably promotes the performance.
Although ours (Basic+NS+CR) is slightly weaker ($\leq0.2\%$ in the final step) than Basic+CR in continual person re-id, ours is more suitable for large-scale continual learning because of its better scalability and efficiency. This priority will be further discussed in Sec. \ref{sec:disc}.

\noindent \textbf{Sensitiveness of Hyper-parameters.}
We further analyze the sensitiveness of performance w.r.t. the two key hyper-parameters, namely, $K$: neighborhood size and $\beta$: margin magnitude, in our method. The experiments are based on 5-step continual person re-id.

\noindent \textbf{Neighborhood Size.} First, we change $K$ in Basic+NS which only includes the neighborhood selection module. The range of $K$ is $\{0, 20, 200, 500, 1000\}$. If the number of old classes in current step is less than $K$, all old classes will be used. As shown in Table \ref{tab:sensitive}, when $K=200$, Basic+NS achieves the best performance. The best $K$ is about $10\%$ of the number of all old classes in the final step. 

\noindent \textbf{Margin Magnitude.} For simplicity, we analyze the performance of Basic+NS +CR with fixed $K$ and varying $\beta$. The range of $\beta$ is $\{0, 2, 5, 10, 50\}\times 10^{-3}$. According to Table \ref{tab:sensitive}, $\beta=10^{-2}$ is the best parameter when $K=200$. Overall, the changing of performance w.r.t $K$ and $\beta$ is smooth.

\begin{table}
\centering
\scriptsize
\setlength{\tabcolsep}{2pt}
\begin{tabular}{cccccc}
\toprule
{$K$}      & {0}              & {20}             & {200}                     & {500}                     & 1000           \\ \hline
Top1                        & $59.1\pm0.1$ & $59.1\pm0.3$        & $\mathbf{59.8\pm0.2}$ & $59.1\pm0.3$          & $58.9\pm0.4$ \\ \hline 
mAP                         & $47.1\pm0.1$ & $47.4\pm0.1$        & $\mathbf{47.8\pm0.1}$ & $47.4\pm0.1$          & $47.0\pm0.2$ \\ \toprule
$\beta$$\times{10^{-3}}$                    & 0              & 2              & 5                       & 10                      & 50             \\ \hline
Top1                        & $59.8\pm0.2$ & $59.7\pm0.3$ & $60.7\pm0.4$          & $\mathbf{61.1\pm0.2}$ & $59.4\pm0.2$ \\ \hline
mAP                         & $47.8\pm0.1$ & $47.9\pm0.2$ & $49.0\pm0.3$          & $\mathbf{49.5\pm0.2}$ & $48.4\pm0.2$ \\ \bottomrule
\end{tabular}
\caption{\footnotesize{The sensitive analysis of performance w.r.t hyper-parameters $K$ and $\beta$. The upper results are based on Basic+NS with varying $K$. The lower results are based on Basic+NS+CR with fixed $K=200$ and varying $\beta$ ($\times{10^{-3}}$). The results are from 5-step continual person re-id.}}
\label{tab:sensitive}
\end{table}

\section{Discussion of Scalability and Efficiency}
\label{sec:disc}
In continual learning, the old classes accumulate quickly along with more learning steps. Especially, in face recognition and person re-id, thousands even millions of identities are involved in large-scale datasets and real applications. 
If we use constant $K$ as the neighborhood size, the memory and time cost for back-propagation of popular methods (LWF, iCaRL and End2End) which use all old classes for knowledge distillation is $\mathcal{O}(t)$ times of ours, and it increases along with step $t$. If we use a constant ratio $\frac{1}{r}$ of old classes, where $r>1$, their memory and time cost is $\mathcal{O}(r)$ times of ours. 
Although we need do feed-forward on all old classes for neighborhood selection, this operation is not time-consuming compared to back-propagation. Besides, feed-forward can be implement on CPU with RAM which is $\times10$ to $\times1000$ larger than GPU memory. 

\section{Conclusion \& Future Work}
\label{sec:conclusion}
In this paper, we propose the continual representation learning for biometric identification with two large-scale benchmarks. Flexible knowledge distillation with Neighborhood Selection and Consistency Relaxation modules are proposed for better scalability and flexibility in large-scale continual learning. Extensive experiments show that our method outperforms the state-of-the-art on two benchmarks. Effectiveness of the two modules is verified by ablation study. 
In the future, more effort should be devoted to improving the generalization ability and scalability of continual learning models in large-scale real-world applications. 


{\small
\bibliographystyle{ieee_fullname}
\bibliography{egbib}
}

\end{document}